\documentclass[aoas]{imsart}

\usepackage{amsthm,amsmath,amsfonts,color}
\usepackage{booktabs}
\usepackage[authoryear]{natbib}
\RequirePackage[colorlinks,citecolor=blue,urlcolor=blue]{hyperref}

\usepackage{graphicx}
\usepackage{multirow}
\usepackage[table,xcdraw]{xcolor}
\usepackage{etoolbox}
\usepackage{caption}
\usepackage{subcaption}

\usepackage{tikz}
\usetikzlibrary{bayesnet}
\usepackage[colorinlistoftodos]{todonotes}
\usepackage{tabulary}

\graphicspath{{./}{figure/}}

\setattribute{journal}{name}{} 

\startlocaldefs

\makeatletter

\patchcmd{\NAT@test}{\else \NAT@nm}{\else \NAT@nmfmt{\NAT@nm}}{}{}

\DeclareRobustCommand\citepos
{\begingroup
 \let\NAT@nmfmt\NAT@posfmt
 \NAT@swafalse\let\NAT@ctype\z@\NAT@partrue
 \@ifstar{\NAT@fulltrue\NAT@citetp}{\NAT@fullfalse\NAT@citetp}}

\let\NAT@orig@nmfmt\NAT@nmfmt
\def\NAT@posfmt#1{\NAT@orig@nmfmt{#1's}}

\makeatother

\newcommand{\reals}{\mathbb{R}}
\newcommand{\Oh}{O}

\newcommand{\T}{\mathrm{T}}

\newcommand{\E}{\operatorname{E}}

\newcommand{\foreign}[1]{\emph{#1}}
\newcommand{\Dail}{\foreign{D\'ail}}
\newcommand{\FiannaFail}{\foreign{Fianna F\'ail}}
\newcommand{\FineGael}{\foreign{Fine Gael}}
\newcommand{\Taoiseach}{\foreign{Taoiseach}}
\newcommand{\Tanaiste}{\foreign{T\'anaiste}}
\newcommand{\TeachtaDala}{\foreign{Teachta D\'ala}}

\newcommand{\Gov}{\emph{Government}}
\newcommand{\Opp}{\emph{Opposition}}

\newcommand{\vocab}{\mathcal{V}}
\newcommand{\word}[1]{\emph{#1}}
\endlocaldefs

\begin{document}

\begin{frontmatter}

 \title{Scaling Text with the Class Affinity Model}
 \runtitle{Scaling Text}

 \begin{aug}
  \author{\fnms{Patrick O.} \snm{Perry}\thanksref{m1}\ead[label=e2]{pperry@stern.nyu.edu}}
  \and
  \author{\fnms{Kenneth} \snm{Benoit}\thanksref{t2,m2}\ead[label=e1]{kbenoit@lse.ac.uk}}

  \thankstext{t2}{This research was supported by the European Research
   Council grant ERC-2011-StG 283794-QUANTESS.}\runauthor{Perry and Benoit}

  \affiliation{New York University\thanksmark{m1} \and London School of
   Economics and Political~Science\thanksmark{m2}}

  \address{Addresses:\\
   Stern School, NYU
   New York, NY 10012;\\
   Department of Methodology, LSE, London WC2A 2AE, UK\\
   \printead{e2}\\
   \phantom{E-mail:\ }\printead*{e1}}
 \end{aug}


 \begin{abstract}

  Probabilistic methods for classifying text form a rich tradition in machine
  learning and natural language processing.  For many important problems,
  however, class prediction is uninteresting because the class is known, and
  instead the focus shifts to estimating latent quantities related to the
  text, such as affect or ideology.  We focus on one such problem of interest,
  estimating the ideological positions of 55 Irish legislators in the 1991
  \Dail\ confidence vote. To solve the \Dail\ scaling problem and others
  like it, we develop a text modeling framework that allows actors to take
  latent positions on a ``gray'' spectrum between ``black'' and ``white'' polar
  opposites.  We are able to validate results from this model by measuring
  the influences exhibited by individual words, and we are able to quantify
  the uncertainty in the scaling estimates by using a sentence-level block
  bootstrap. Applying our method to the \Dail\ debate, we are able to
  scale the legislators between extreme pro-government and pro-opposition in a
  way that reveals nuances in their speeches not captured by their votes or
  party affiliations.

 \end{abstract}



\end{frontmatter}

\section{Introduction}

\noindent Text classification, where the goal is to infer a discrete class
label from observed text, is a core activity in statistical and machine
learning and natural language processing.  Instances of this problem include
inferring authorship \citep{MostellerWallace1963} or genre
\cite{Kessler+1997}, detecting deception \citep{Newman+2003}, classifying
e-mail as ``spam'' \citep{Heckerman+1998}, or detecting
sentiment \citep{Pang+2002}. The huge appeal of the methods developed for
these applications is that, from a small training set, it is possible to
classify a large number of unlabelled documents to reasonable accuracy without
costly human intervention.

In many applications, however, classification is an uninteresting goal, since
the correct identification of the class is obvious and costless.  It is
fundamentally uninteresting, for example, to attempt to predict the political
party of a speaker or the identity of a Supreme Court justice.  Furthermore,
in many social and political settings with observed discrete outcomes,
institutions may cause predicted and observed class membership to diverge in
significant ways.  In parliamentary democracies where party discipline is
enforced, for instance, voting may follow party lines even if the best
predictions from observable features indicate more heterogeneous outcomes.  In
such cases, it is trivial to predict class (a legislator's vote) from
observable covariates (political party). In the presence of these covariates,
the text of a speech is ancillary to the goal of class label prediction.

Even when observing text does not improve prediction performance, it is not
the case that text is uninformative.  In legislative debates, the text that
legislators generate through floor speeches may provide a direct opportunity
for them to express their contrary and divergent preferences \cite[see for
instance][]{BenoitHerzog2012}.  With legal briefs, to take
another example, it is trivial to classify opinions as majority or dissenting
but using the observed text and other information it is possible to place the
briefs on a spectrum between the two extremes
\citep{ClarkLauderdale2010}.  Simply attempting to predict the
category of opinion---for instance classifying \emph{amicus curiae} briefs as
pro-petitioner or pro-respondent \cite[e.g.][]{Evans+2007}, is of less direct
interest since these categories are already known. The text of a document can
reveal nuances that are not captured by and sometimes in disagreement with its
class label.

\begin{table} 
 \caption{Irish D\'ail debate speech statistics.}\label{tab:speech-stats}
 \begin{tabular}{llcll}
  \toprule
  \multicolumn{2}{l}{\emph{Government party members}}
  && \multicolumn{2}{l}{\emph{Opposition party members}} \\
  Fianna F\'ail (FF) & 24
  && Democratic Left (DL) & \phantom{0}3 \\
  Progressive Dems. (PD) & \phantom{0}1
  && Fine Gael (FG) & 22 \\
    &   &   & Green        & \phantom{0}1 \\
    &   &   & Labour (Lab) & \phantom{0}7 \\
  \\
  \emph{Speech text} \\
  Median length (leaders)
  & \multicolumn{3}{l}{6,348 tokens} \\
  Median length (others)
  & \multicolumn{3}{l}{2,210 tokens} \\
  Vocabulary size
  & \multicolumn{3}{l}{9,731 word types} \\
  \bottomrule
 \end{tabular}
\end{table}

Here, we focus on an application that is ill-suited to
text classification but where text is nonetheless informative.  We analyze the
1991 Irish \Dail\ confidence debate, previously studied by
\citet{LaverBenoit2002} who used the debate speeches to demonstrate their
``Wordscores'' scaling method.  The context is that in 1991, as the country
was coming out of a recession, a series of corruption scandals surfaced
involving improper property deals made between the government and certain
private companies.  The public backlash precipitated a confidence vote in the
government, on which the legislators (each called a \TeachtaDala, or TD) debated and then voted to decide whether
the current government would remain or be forced constitutionally to resign.
Table~\ref{tab:speech-stats} summarizes the composition of the \Dail\ in 1991
and provides some descriptive statistics about the speech texts.  We can use
the debate as a chance to learn the legislators' ideological positions.

Because the Irish parliamentary context is characterized by strict party
discipline, the move was largely symbolic and each legislator voted strictly
with his or her party: all members of the governing parties (\FiannaFail\ and
the Progressive Democrats) voted to support the government, and all members of
the opposition parties (the Democratic Left, \FineGael, Green, and Labour)
voted against.

Despite the votes being entirely predictable, the floor speeches from the
debate before the official tally reveal nuances to legislators' positions.
Take, for example, the following excerpt
from Noel Davern, a moderate from the \FiannaFail\ party:
\begin{quotation}
 \noindent
 It is not that the financial scandals have not occurred. They have occurred and the Government have taken quick action on them. In fact, we are not fully qualified to speak on them until we see the results of the full and independent inquiry.
\end{quotation}
Davern supports the government, but at the same time does
not excuse them from all culpability.
Contrast this with a typical opposition speech, calling for a vote against the confidence motion, from Labour TD Michael Ferris:
\begin{quotation}
 \noindent
 Our decision to oppose this motion of confidence is a positive assertion of the disapproval of the ordinary people of the actions of this discredited Government. The people have watched with amazement the unfolding of scandals which have tainted this Government. The Government cannot now be said to deserve the confidence of the people.
\end{quotation}
Both legislators express views that place them somewhere between the two
extremes of absolute government support and absolute opposition support.

Where do Davern, Ferris, and the other 56 TDs that participated
in the debate lie on this ideological spectrum? This is the essential question
that we attack in this manuscript. In answering the question, we have at our
disposal the speech texts, along with some additional information. We know
that the leader of the government (Haughey, the \FiannaFail\ \Taoiseach)
will give a speech at one extreme of the pro-government
spectrum, and we know that the heads of the two major opposition parties
(Spring and De~Rossa, the Labour the Democratic Left leaders) will be at the
extreme of the other end. We will use these three texts as reference points
by which to scale the other 55 ambiguous texts whose positions are unknown
and must be estimated.


To solve our particular problem, we develop a new text scaling method
that is broadly applicable to situations where most documents are unlabelled
but we have a few examples of documents at the extremes of a hypothesized
ideological or stylistic spectrum.  Instead of predicting class membership,
our objective in such problems is to \emph{scale a continuous characteristic},
through measuring the fit of a text to a set of known classes based on its
degree of similarity to typical texts from these classes.

In what follows, we develop the \emph{class affinity model} and demonstrate
its use in
scaling the degree of support or opposition expressed in the speeches made
during the confidence debate.  We start by outlining the foundations of our
scaling
model, contrasting it first to similar approaches designed for classification
(Section \ref{sec:scalingwithclassification}), and then to lexicographical
association methods in the form of sentiment dictionaries (Section
\ref{sec:dictionaries}).  Section \ref{sec:model} then sets out the model,
comparing this to related methods, highlighting the differences through on
statistical principles but also using our application.   Sections
\ref{sec:estimating-affinities} and \ref{sec:estimating-ref-dists} detail how
this model and its reference distributions are estimated, while Section
\ref{sec:othermethods} relates the affinity model to related methods.  In
Section \ref{sec:diagnostics}, we show how to measure the influence of
individual words, and provide recommendations for removing common terms that
might skew the results.  We apply this procedure to choose a tailored
vocabulary for our application in Section~\ref{sec:vocabulary}.
Section \ref{sec:uncertainty quantification}
demonstrates how to estimate uncertainty for the class affinity scaled
estimates.  Finally, we summarize the results the results of fitting the class
affinity model to our application (Section \ref{sec:results}), and offer some
concluding remarks.

\section{Scaling with a classification method}
\label{sec:scalingwithclassification}

We have stated repeatedly that classification is not our objective in this
problem, but nonetheless there is a long tradition of fitting classification
methods to text, and we might try applying one of those methods here.  We have
a ``training set'' of the three leadership speeches, one of which we can label
as \Gov\ and two as \Opp. We can fit a supervised classification method to
this training set and then use it to make predictions for the other 55
legislators.

Using the Naive Bayes text classification method popularized by
\citet{Sahami+:1998}, we would model the tokens in each speech text as
independent draws from a label-dependent distribution estimated from the
reference texts. Letting label $k = 1$ denote \Gov\ and label $k =
2$ denote \Opp, for each label $k \in \{ 1, 2 \}$ and word type
$v$ in our vocabulary $\vocab$, we would estimate $p_{kv}$, the probability
that a random token drawn from a text with label $k$ is equal to $v$.
Typically we use the empirical word occurrence frequencies in the reference
documents or some smoothed version thereof.  Here and throughout the text,
unless otherwise noted we will take our vocabulary to be the set of word types
that appear at least twice in the leadership speeches, excluding common
function words from the modified Snowball stop word list distributed with the
\texttt{quanteda} software package \citep{Porter06, quanteda}; we ignore words
outside this set.

Under the ``naive'' assumption
that tokens in a text are independent draws from the same distribution,
assuming equal prior odds for each label, the log-odds that the label is
\Gov\ given the word counts $x = (x_v)_{v \in \vocab}$ is
\[
 \eta(x) = \sum_{v \in \vocab} x_v \log(p_{1v}/p_{2v}),
\]
where $x_v$ denotes the number of times that word type $v$ appears in the text.
The expression for $\eta(x)$ arises as the log ratio of two multinomial
likelihoods with probability vectors $p_1$ and $p_2$. Using Naive Bayes
classification for this two-class prediction problem,
we would predict the label as \Gov\ when
$\eta(x) > 0$, and we would predict the label as \Opp\ when
$\eta(x) < 0$.

\begin{figure}
 \includegraphics[height=3in]{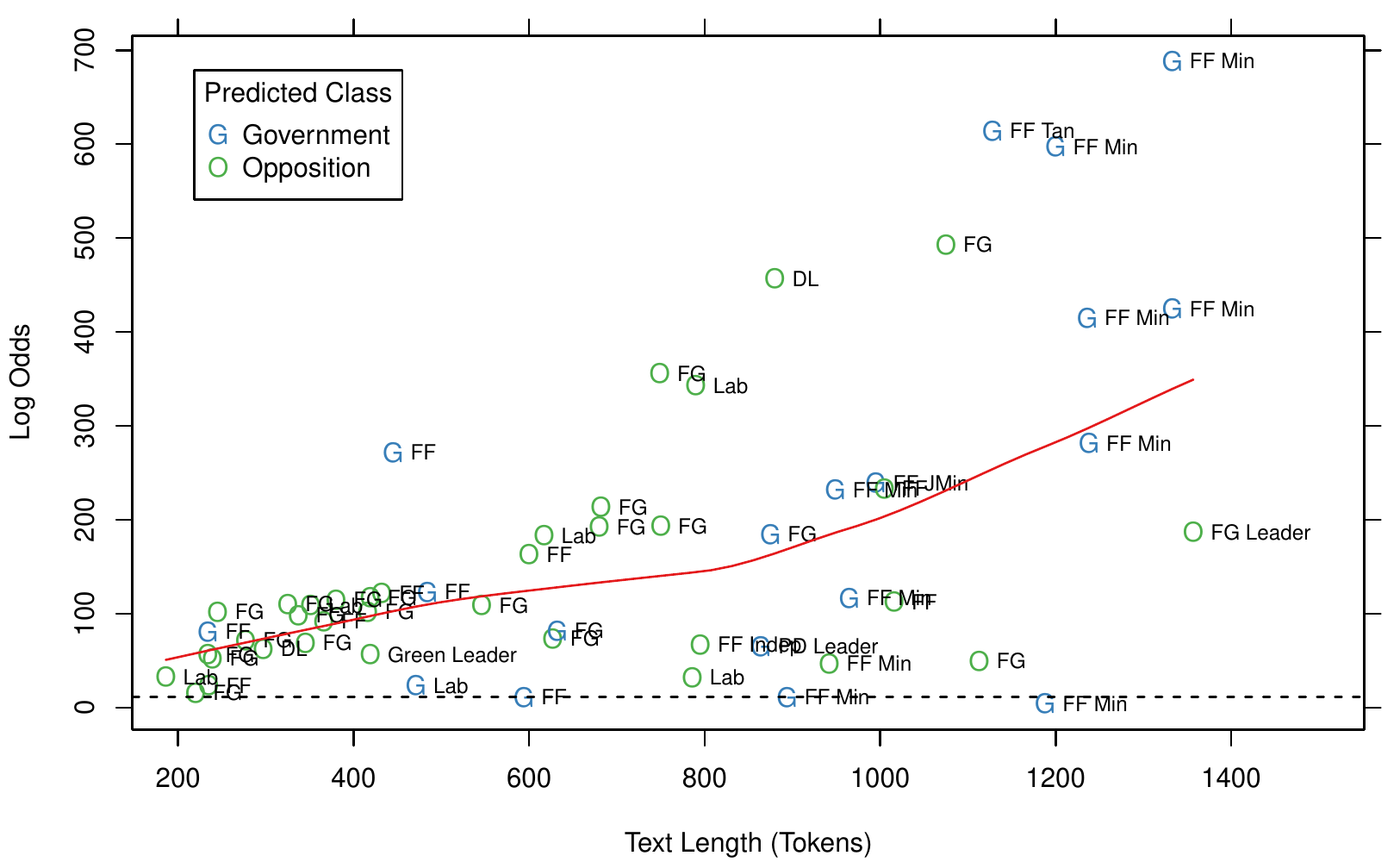}
 \caption{Odds of class membership for the debate speeches as
  predicted by a Naive Bayes model.
  Points above the dashed lines have predicted class probabilities exceeding
  99.999\%.}\label{fig:naive_bayes-len}
\end{figure}

The quantity $\eta(x)$ measures the strength of the evidence that the label of
a text is \Gov\ or \Opp, and we can use this quantity
to scale the 55~virgin texts.
Unfortunately, the Naive Bayes scaling method has serious drawbacks.  First,
the estimated log odds tend to be absurdly high. On our example, the median
absolute log odds is $197.8$, corresponding to an unrealistically high
probability of class membership exceeding $1 - 10^{-85}$.
Second, because $\eta(x)$ is measuring the strength of the
evidence, longer texts will tend to have higher absolute log odds. We
illustrate both of these defects in Fig.~\ref{fig:naive_bayes-len}, where we
plot the absolute odds of class membership as a function of text length.

Related methods suffer from versions of this same problem.  Multinomial
inverse regression \citep{Taddy2013} regularizes the probability vector
estimates $p_1$ and $p_2$ adds a calibration step to the log-odds, but it
still suffers from the same drawbacks as Naive Bayes.  Discriminative methods,
like those used by \citet{Joachims1998} and 
\citet{Jia+2014}, are affected to a degree depending on their choice of
features.  With logistic regression, for example, when the features are linear
functions of the counts $x$, then it will still be the case that longer
documents have more extreme counts and hence more extreme predictions. Other
choices of predictors can give rise to predictors that are less sensitive to
variations in document length.

Even if these classification methods did not suffer the defects noted above,
there is still a fundamental disconnect between the classification philosophy
and the goals of scaling.  In the classification world, a document is either
``black'' or ``white;'' for an unlabelled document, the method will tell you
the probability that the label is black. In reality, though, a text is
``gray,'' a mixture of black and white.  This is a fundamental difference in
perspective that precludes using a classification method for our task. We
expand on this metaphor below.

\section{Scaling with dictionaries}
\label{sec:dictionaries}

Not all text scaling methods take the black-and-white classification view of
the world. One of the most successful alternatives is dictionary-based scaling
\citep{Stone+1966,Pennebaker+2001,HuLiu2004}. In their simplest forms,
dictionary methods conceive as each text as a mixture of two contrasting
poles, such as positive and negative. Neutral words get discarded from the
vocabulary. The scaling of a text is determined by the average orientation
of its tokens.

There are many variations of dictionary-based scaling 
but for concreteness we will focus on \citepos{GrimmerStewart2013}
formulation. To
apply that scaling to the problem at hand---scaling debate speeches---we would
need two non-overlapping lists: one of words associated with 
\Gov\ and one of words associated with \Opp.  Given these lists, we
would assign a score $s_v = +1$ to each word type $v$ in the
\Gov\ list, and a score $s_v = -1$ to each word type $v$ in
the \Opp\ list. The dictionary-based scaling of a text with token
count vector $x$ would be
\[
 t(x)
 =
 \frac{1}{n} \sum_{v \in \vocab} x_v \, s_v,
\]
where $n = \sum_{v \in \vocab} x_v$; this quantity is equal to the
difference in word type occurrence rates between the \Gov\ and
\Opp\ lists.


It is labor-intensive and error-prone to build a custom dictionary for
each application, so
often when practitioners apply dictionary scaling methods, they use
off-the-shelf dictionaries instead of building their own. For our application,
the Lexicoder sentiment dictionary (LSD, 2015 version), ``a broad lexicon
scored for positive and negative tone and tailored primarily to political
texts,'' would be a natural choice \citep[211]{youngsoroka2012}.  However, as
those authors note, applying an
off-the-shelf dictionary to a new domain often leads to undesirable results.
Table~\ref{tab:dictionary-v-lsd} illustrates this point in the context of our
application by comparing the word orientations as determined by the LSD with
their empirical associations with \Gov\ and \Opp\ as
observed in the leadership speeches. The rows indicate the LSD-assigned
orientations of the words; the columns are the significant differences in
usage rates between the two classes as measured by the ``keyness'' $G^2$
likelihood ratio score at significance level $0.05$, taking negations into
account as
recommended by \citet{youngsoroka2012}. We display the number of word types in
each cell, along with the most common words.

\begin{table}
 \centering
 \caption{Comparing government and opposition words to Lexicoder sentiment dictionary matches.}
 \label{tab:dictionary-v-lsd}
 \scriptsize
 \begin{tabulary}{\textwidth}{lCCC}
  \toprule
  & \multicolumn{3}{c}{\textit{Government/Opposition}}                                                                                                                                                                                                                                                                                                    \\
  \textit{Sentiment} & Government
   & Neutral
   & Opposition
   \\ \midrule
  Positive           & 11 & 377 & 2 \\
  & partners, progress, balance, achieved, legitimate, best, forward, better,
   improvement, improvements             &  confidence, like, great, well,
   ensure, hope        good, opportunity, normal, responsible           &
   wealth, creation             \\ \midrule
  Neutral            & 66 & 2,329 & 54 \\
  &  public, now, economic, per, economy, cent, growth, new, way, community
   & government, country, business, irish, made, many, us, can, years, must
   &  people, political, house, mr, one, taoiseach, minister, deputy, time,
   questions \\ \midrule
  Negative           & 8 & 346 & 0 \\
  &  problems, ireland's, debt, difficulties, deficit, deterioration,
   opposite, implications & scandals, ireland, difficult, allegations, failed,
   concern, scandal, unfortunately, innuendo, loss &  ---  \\ \bottomrule
 \end{tabulary}
\end{table}

If the dictionary were appropriate for our application, we should observe
positive words associated with government usage, and negative words associated
with opposition usage.  The patterns in Table~\ref{tab:dictionary-v-lsd},
however, show a very different result. Only 11 ``positive'' words have high
usage in the government leadership speech, and no ``negative'' words
have high usage in the opposition leadership speeches. Most
``positive'' and ``negative'' words do not have a clear association with
either \Gov\ or \Opp. Furthermore, there are some
worrying cases where the dictionary orientation is counter to the association
between the classes. For example, while the LSD declares the word to be
negative, in the
context of the debate \word{deficit} refers simply to a fiscal outcome;
likewise, \word{confidence} is related to the question of the debate, and not
intended to convey positive valence.
Despite being designed to detect political valence,
the dictionary fails here since it has not been tailored for this
particular debate. Terms that are associated with one type of affect
generally are used differently in the context of the no-confidence debate.



Beyond the problem of domain adaptation, the more fundamental
issue with dictionary methods is that their basic premise---that each word has
a clear orientation---is inappropriate in our domain.  Most words in our
application do not clearly either belong in one category or the other.  We can
seen this in Table~\ref{tab:dictionary-v-lsd}, where over~95\% of the word
types do not have statistically significantly different usage
rates between the government and opposition leadership speeches.
The vast majority of words get used by both government and opposition, and
thus have mixed associations with both classes.  Some dictionaries try to
adjust for this by giving non-binary scores to the words
\citep{BradleyLang:1990}, but these adjustments are often \emph{ad hoc}, and
they suffer from the same domain adaption problems. In the sequel, we present an
alternative method that allows for mixed word association while simultaneously
adapting to the domain.

\section{The affinity model}
\label{sec:model}

Classification methods assume that each text is a member a well-defined
category.  Dictionary methods do not make this strong assumption, but they too
take an unrealistic view of the world by supposing that each word has a
well-defined orientation. Table~\ref{tab:assumptions} highlights this
difference, and makes clear that there is room for a third worldview allowing
both texts and words to be gray. We will formalize this intuition in a
statistical model that we refer to as the ``affinity model.''

\begin{table}[]
 \centering
 \caption{Word- and document-level assumptions from three scaling methods.}
 \label{tab:assumptions}
 \begin{tabular}{lccc}
  &  & \multicolumn{2}{c}{Documents} \\
    &   & \textit{Gray} & \textit{B/W} \\
  \cline{3-4}
  & \multicolumn{1}{c|}{\textit{Gray}} & \multicolumn{1}{c|}{Affinity Model}
  & \multicolumn{1}{c|}{Classification} \\[3ex]
  \cline{3-4}
  \multirow{-2}{*}{Words} & \multicolumn{1}{c|}{\textit{B/W}}
  & \multicolumn{1}{c|}{Dictionaries}
  & \multicolumn{1}{c|}{\cellcolor[HTML]{C0C0C0}} \\[3ex]
  \cline{3-4}
 \end{tabular}
\end{table}

Our basic conceptual model is that over the course of a speech, a speaker's
orientation switches back and forth between \Gov~mode and
\Opp~mode. When she is in \Gov~mode, she chooses
words in the same manner as the government leadership. Likewise, when she is
\Opp~mode, she chooses words in the same manner as the opposition
leadership. We should place the speaker on the spectrum between the two
extremes of pro-government and pro-opposition according to what proportion of
time she spends in each mode.

Formally, let $\vocab$ denote the vocabulary of word types, a set with
cardinality $|\vocab| = V$. Encode the text of a speech as a sequence of
tokens $W = (W_1, W_2, \dotsc, W_n)$, with each token $W_i$ belonging to
$\vocab$.  In our model, the speaker's underlying orientation evolves in
parallel to the text and can be represented as $U = (U_1, U_2, \dotsc, U_n)$
where for $i = 1, \dotsc, n$ the value $U_i$ denotes the speaker's underlying
orientation while uttering token $W_i$. We will in general suppose that there
are $K$ possible orientations, identified with the labels $1, \dotsc, K$.

In our conceptual framework, a speech and the corresponding underlying
orientation sequence are realizations of some speaker-specific random process.
For $k = 1, \dotsc, K$, we define a speaker's affinity toward orientation~$k$
as $\theta_k$, the expected proportion of time that her underlying orientation
is~$k$:
\[
 \theta_k = \E\Big\{\frac{1}{n} \sum_{i=1}^{n} U_i \Big\}.
\]
Each speaker has an underlying affinity vector
$\theta = (\theta_1, \dotsc, \theta_K)$.

In our specific application, there are $K = 2$ orientations.  Each debate
speaker has a separate affinity vector $\theta = (\theta_1, \theta_2)$.  We
will scale each speaker by estimating his or her affinities for
\Gov~($\theta_1$) and \Opp~($\theta_2$).

\begin{figure}
 \begin{center}
  \begin{tikzpicture}
   \node[align = center, text width=4cm] (labtheta) at (0,1) {Speaker affinity};
   \node[align = center, text width=4cm, below=of labtheta] (labU) {Intended class};
   \node[align = center, text width = 4cm, below=of labU, yshift = -.2cm] (labW) {Observed words};

   \node[const, left=of labtheta, xshift = -.5cm] (t) {$\theta$};
   \node[latent, below=of t, xshift = -1cm] (u1) {$U_1$};
   \node[latent, below=of t, xshift = 0cm] (u2) {$U_2$};
   \node[const, below=of t, xshift = .75cm, yshift = -.3cm] (udots) {$\cdots$};
   \node[latent, below=of t, xshift = +1.5cm] (un) {$U_n$};
   \node[obs, below=of u1] (w1) {$W_1$};
   \node[obs, below=of u2] (w2) {$W_2$};
   \node[const, below=of udots, yshift = -.6cm] (wdots) {$\cdots$};
   \node[obs, below=of un] (wn) {$W_n$};
   \edge[->, shorten <= 5pt] {t} {u1,u2,un};
   \edge[->] {u1} {w1};
   \edge[->] {u2} {w2};
   \edge[->] {un} {wn};

   \node[const, right=of labtheta, xshift = 0cm] (t2) {$\theta$};
   \node[latent, below=of t2] (u_2) {$U$};
   \node[obs, below=of u_2, xshift = -1cm] (w1_2) {$W_1$};
   \node[obs, below=of u_2, xshift = -0cm] (w2_2) {$W_2$};
   \node[const, below=of u_2, xshift = .75cm, yshift = -.3cm] (wdots_2) {$\cdots$};
   \node[obs, below=of u_2, xshift = +1.5cm] (wn_2) {$W_n$};
   \edge[->, shorten <= 5pt] {t2} {u_2};
   \edge[->] {u_2} {w1_2};
   \edge[->] {u_2} {w2_2};
   \edge[->] {u_2} {wn_2};

   \node[below=of w2, xshift = .25cm, yshift = 0.6cm] (caption1) {(a) Class affinity model};
   \node[below=of w2_2, xshift = 0.25cm, yshift = 0.6cm] (caption2) {(b) Classification Model};
  \end{tikzpicture}
 \end{center}
 \caption{Generative model for the underlying orientation $U$
 and the token sequence $W$, contrasting the class affinity model to the classification model.}
 \label{fig:generative}
\end{figure}
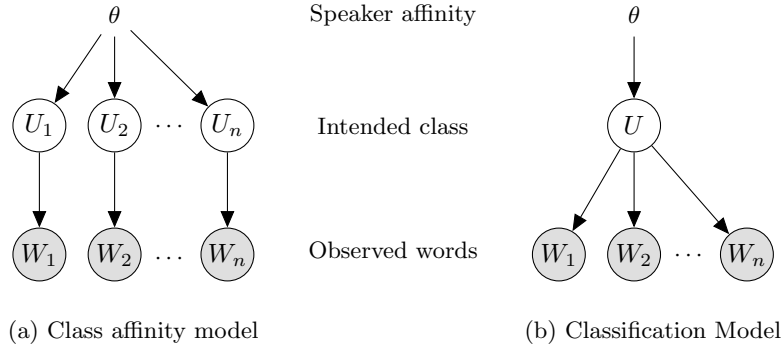

We will impose two simplifying assumptions to make inference under our model
tractable. First, we will suppose that $U_1, U_2, \dotsc, U_n$ are independent
and identically distributed. This forces that for every label~$k$, and
position~$i$, the underlying orientation is randomly distributed with
$\Pr(U_i = k) = \theta_k$.
Second, we will suppose that $W_1, W_2, \dotsc,
W_n$ are independent conditional on $U$, and that the distribution of
$W_i \mid U$ depends only on $U_i$ and is the same for all positions $i$.
This positional invariance allows us to define
for each label $k$ and word type $v$ the probability
\[
 p_{kv} = \Pr(W_i = v \mid U_i = k)
\]
and it allows us to define the reference distribution $p_k = (p_{kv})_{v \in
 \vocab}$. Our two simplifying assumptions result in a generative model: for
each position $i = 1, \dotsc, n$, the speaker picks an underlying orientation
with probabilities determined by $\theta$; given that the underlying
orientation is $U_i = k$, the speaker picks token $W_i$ according to
distribution $p_k$.  Fig.~\ref{fig:generative}(a) summarizes this generative
process.

For each position $i = 1, \dotsc, n$, the chance that word
$v$ appears in position~$i$ is
\begin{align*}
 \Pr(W_i = v)
   & = \sum_{k=1}^{K} \Pr(U_i = k) \Pr(W_i = v \mid U_i = k)
    = \sum_{k=1}^{K} \theta_k \, p_{kv}.
\end{align*}
Further, $W_1, W_2, \dotsc, W_n$ are independent, so that the probability
of observing the token sequence
$w = (w_1, \dotsc, w_n)$ is
\begin{equation}\label{eqn:affinity-lik}
 \Pr(W = w)
 = \prod_{i=1}^{n}
 \big(\sum_{k=1}^{K} \theta_k \, p_{kw_i}\big)
 = \prod_{v \in \vocab}
 \big(\sum_{k=1}^{K} \theta_k \, p_{kv}\big)^{x_v},
\end{equation}
where $x_v$ is the number of times word~$v$ appears in the text.
At a high level, this is the same generative model as that used for a
topic model \citep{Blei+2003}.  The main difference
between these models is that topic models are typically unsupervised, but
the affinity model uses supervision to estimate $p_1, p_2, \dotsc, p_K$.
We elaborate more on the connection to topic models in
Section~\ref{sec:topic-models}.

We note also that the affinity model can be seen as a generalization of the
Naive Bayes model depicted in Fig.~\ref{fig:generative}(b). In the Naive
Bayes model, each document has a single underlying orientation, $U$. All words
in the document share the same underlying orientation. The parameter $\theta$
can be seen as the prior distribution for $U$. In Naive Bayes, we do not
estimate $\theta$, but instead we estimate $\Pr(U = k \mid X_1, \dotsc, X_n)$
for each class $k$. In Naive Bayes, each document has just one underlying
orientation.  The power of the affinity model is that it allows the underlying
orientation to vary with the word position.

\section{Estimating affinities}
\label{sec:estimating-affinities}

The affinity model described in Section~\ref{sec:model} lends itself naturally
to likelihood-based estimation. We first consider the problem of estimating
the affinity vector~$\theta$ for a particular text, when we are given the
reference distributions $p_1, \dotsc, p_K$.

The parameter space for the affinity vector is the simplex $\Theta \subset
\reals^{K}$ consisting of all vectors $\theta$ with non-negative components
satisfying the equality constraint $\sum_{k=1}^{K} \theta_k = 1$.
One implication of the equality constraint is that the model
is over-parametrized, which makes estimating $\theta$ directly awkward.
To handle this constraint, we will
reparametrize the model in terms of a $(K - 1)$-dimensional contrast
vector~$\beta$.

In the $K = 2$ case, we set
\(
\beta = (\theta_2 - \theta_1) / 2,
\)
so that
\(
\theta_1 = 1/2 - \beta
\)
and
\(
\theta_2 = 1/2 + \beta;
\)
the parameter space for $\beta$ is $\mathcal{B} = [-1/2,1/2]$. In the
general case we let $\beta$ be defined by the relation
\begin{equation}\label{eqn:theta-beta}
 \theta = \theta_0 + C \beta,
\end{equation}
where $\theta_0$ is any point in the interior of the parameter space and
the contrast matrix $C \in \reals^{K \times (K-1)}$ has full rank and
satisfies $C^\T 1 = 0$.  In principle $\theta_0$ and $C$ can be arbitrary, but
for concreteness we will take $\theta_0$ to be the center of the parameter
space $\theta_0 = (1/K, 1/K, \dotsc, 1/K)$, and we will take $C$ to be the
Helmert matrix. 
The parameter space for the contrast vector, then, is
\(
\mathcal{B} = \{ \beta \in \reals^{K-1} :
\theta_0 + C \beta \succeq 0 \},
\)
where $\succeq$ denotes component-wise partial order.
With this particular choice of $\theta_0$ and $C$, the general case agrees
with the special case when $K = 2$.

Following equation~\eqref{eqn:affinity-lik}, the log-likelihood function
for the contrast vector is
\begin{equation}\label{eqn:affinity-loglik}
 l(\beta)
 =
 \sum_{v \in \vocab}
 x_v
 \log \mu_v,
\end{equation}
where
\(
\mu_v = \sum_{k=1}^{K} \theta_k \, p_{kv}
\)
and
\(
\theta = \theta(\beta).
\)
We will estimate $\beta$ by maximizing $l(\beta)$ or a penalized version
thereof.

In the special case when $K = 2$, the score and observed information
functions gotten from differentiating the log likelihood are
\begin{align*}
 u(\beta)
   & = l'(\beta)
 = \sum_{v \in \vocab} \frac{p_{2v} - p_{1v}}{\mu_v} x_v,
 \\
 I(\beta)
   & =
 -l''(\beta)
 =
 \sum_{v \in \vocab} \frac{(p_{2v} - p_{1v})^2}{\mu_v^2} x_v.
\end{align*}
The expected information is
\begin{equation*}
 i(\beta) = \E\{ I(\beta) \}
 =
 n
 \sum_{v \in \vocab}
 \frac{(p_{2v} - p_{1v})^2}{\mu_v}.
\end{equation*}

To define the analogous functions in the general case, define the
matrix-valued function
\(
Q = Q(\beta) \in \reals^{K \times \vocab}
\)
with $Q_{kv} = p_{kv} / \mu_v$.
In the general case, the analogous functions are
\begin{align}
 u(\beta)
   & = C^\T Q x,        \\
   \label{eqn:affinity-oinfo}
 I(\beta)
   & = C^\T Q X Q^\T C,
\end{align}
where $X \in \reals^{\vocab \times \vocab}$ is the diagonal matrix
with $X_{vv} = x_v$ for $v \in \vocab$.
The expected information is
\[
 i(\beta)
 = n \, C^\T Q P^\T C
 = n \, C^\T P Q^\T C,
\]
where $P \in \reals^{K \times \vocab}$ is the matrix with $k$th row
equal to $p_{k}^\T$ for $k = 1, \dotsc, K$.

The observed information function $I(\beta)$ is positive semidefinite,
indicating that the log likelihood function $l(\beta)$ is concave. We can
estimate $\beta$ by maximizing the log likelihood using the Newton-Raphson
iterative method. The expensive part of this maximization procedure is
computing $I(\beta)$, which takes time $\Oh(V K^2),$ or faster if the count
vector $x$ is sparse. In our experience on the \Dail\ speeches, the method
typically converges after about five iterations. The difficult part of the
optimization is that we must restrict the search to the parameter space
$\mathcal{B}$; we accomplish this using an interior-point barrier
method \citep[Ch.~11]{BoydVandenberghe2004}.

In exchange for adding a small bias to the estimates, we can reduce the
variance and remove the explicit inequality constraints on the parameter
space. In particular, \citet{Firth1993} shows that in the
asymptotic regime where $n$ tends to infinity, adding a penalty of
order~$\Oh(1)$ to a log likelihood adds a term of size $\Oh(1/n)$ to the
bias of the estimator (sometimes reducing the estimator's bias, but not
necessarily doing so in our setting). In our case, we choose a positive
scalar $\lambda$ and define the penalty function
\[
 \psi_\lambda(\theta)
 =
 \lambda
 \sum_{k=1}^{K}
 \log \theta_k.
\]
Then, we estimate the affinities by maximizing the penalized log likelihood
\(
 \tilde l_\lambda(\beta)
 =
 l(\beta)
 +
 \psi_\lambda(\theta),
\)
where $\theta = \theta(\beta)$.  The penalty ensures that
$\tilde l_\lambda$ is strictly concave, and further that the
maximizer~$\hat \beta_{\lambda}$ is unique and belongs to the interior
of the parameter space. For the analyses in this manuscript, we use the
penalty value $\lambda = 0.5$. Section~\ref{sec:estimating-ref-dists} provides
some theoretical justification for this penalty value in a related context.

\section{Estimating reference distributions}
\label{sec:estimating-ref-dists}

The reference distributions $p_1, p_2, \dotsc, p_K$ themselves need to be
estimated from data.  In our framework, this learning step requires not
large volumes of training data, but rather texts that are clearly polar examples
of  each reference class, to form benchmarks for estimating the other texts'
affinities to these classes. In the context of our specific application,
the 1991 Irish~\Dail~confidence debate, recall that
the contrasting $K = 2$ classes represent \Gov~($k = 1$) and
\Opp~($k = 2$). We will use the leaders of the government and opposition
respectively
to represent the archetype texts for each class. \Taoiseach\ (Prime Minister)
Charles Haughey's speech forms the government reference text for estimating
$p_1$, and the speeches from the two opposition party leaders (Spring and de
Rossa) form the reference texts for estimating $p_2$.

To estimate a particular reference distribution $p$, we will suppose in
general that we
have at our disposal $m$ texts drawn from this distribution of lengths $n_1,
n_2, \dotsc, n_m$. We denote the vectors of word counts for these texts by
$x_1, x_2, \dotsc, x_m$. In our application, $m = 1$ for estimating the
\Gov~reference, and $m = 2$ for estimating the \Opp~reference. We will use
smoothed empirical frequencies to estimate $p_v$ as
advocated by \citet{Lidstone1920}. We choose a nonnegative smoothing constant
$\alpha$ and estimate the probability of word type $v$ as
\[
 \hat p_v
 =
 \Big(\alpha + \sum_{j=1}^{m} x_{jv}\Big)
 \Big/
 \Big(V \alpha + \sum_{j=1}^{m} n_j\Big).
\]
Specifically, we will set $\alpha = 0.5$.
It is not essential to smooth the estimates of $p$, but doing so reduces
estimation variability.

There are many reasonable choices for the smoothing constant $\alpha$,
including choosing $\alpha$ adaptively \citep{FienbergHolland1972}.  In
natural language processing, it is common to take $\alpha = 1$ so that $\hat
p$ is the maximum \emph{a posteriori} estimator under a uniform prior
\citep[Sec.~4.5.1]{JurafskyMartin2009}. From a frequentist standpoint, the
value $\alpha = 0.5$---which corresponds to using a Jeffreys prior
for~$p$---is slightly more defensible. In the regime where $V$ is fixed and
$n$ tends to infinity, using the results from \citet{Firth1993} one can show
that using $\alpha = 0.5$ results in an expected Kullback-Leibler divergence
from $\hat p$ to $p$ of order $\Oh(n^{-3/2})$ instead of $\Oh(n^{-1})$ for
other choices of $\alpha$.

Once we have estimates $\hat p_1, \hat p_2, \dotsc, \hat p_K$ of the reference
distributions, to get an estimate of the class affinity vector~$\theta$ for a
text, we use the methods from Section~\ref{sec:estimating-affinities}, using
the estimated class distributions in place of their true values.

\section{Connections to other methods}
\label{sec:othermethods}

\begin{figure}
 \centering
 \begin{subfigure}[b]{0.425\textwidth}
  \includegraphics[width=\textwidth]{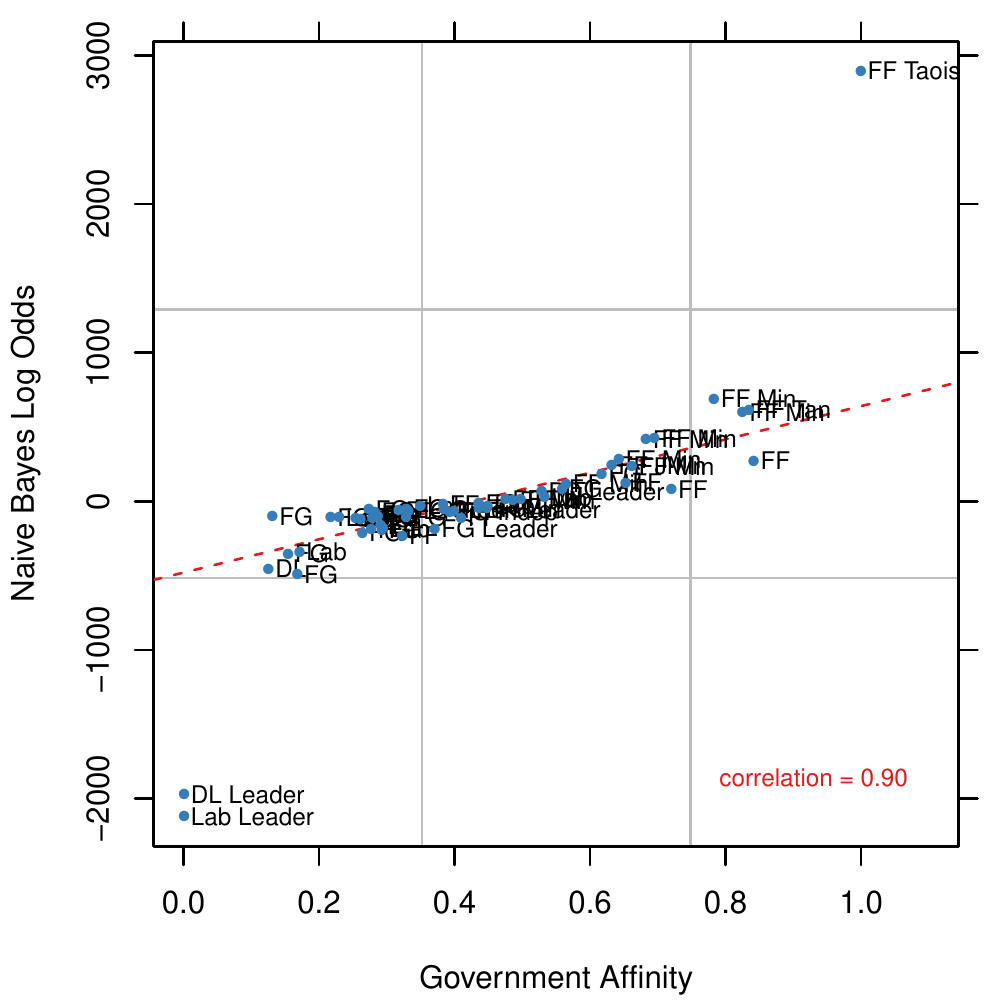}
  \caption{Naive Bayes}
  \label{subfig:naive_bayes-corr}
 \end{subfigure}
 ~\qquad
 \begin{subfigure}[b]{0.425\textwidth}
  \includegraphics[width=\textwidth]{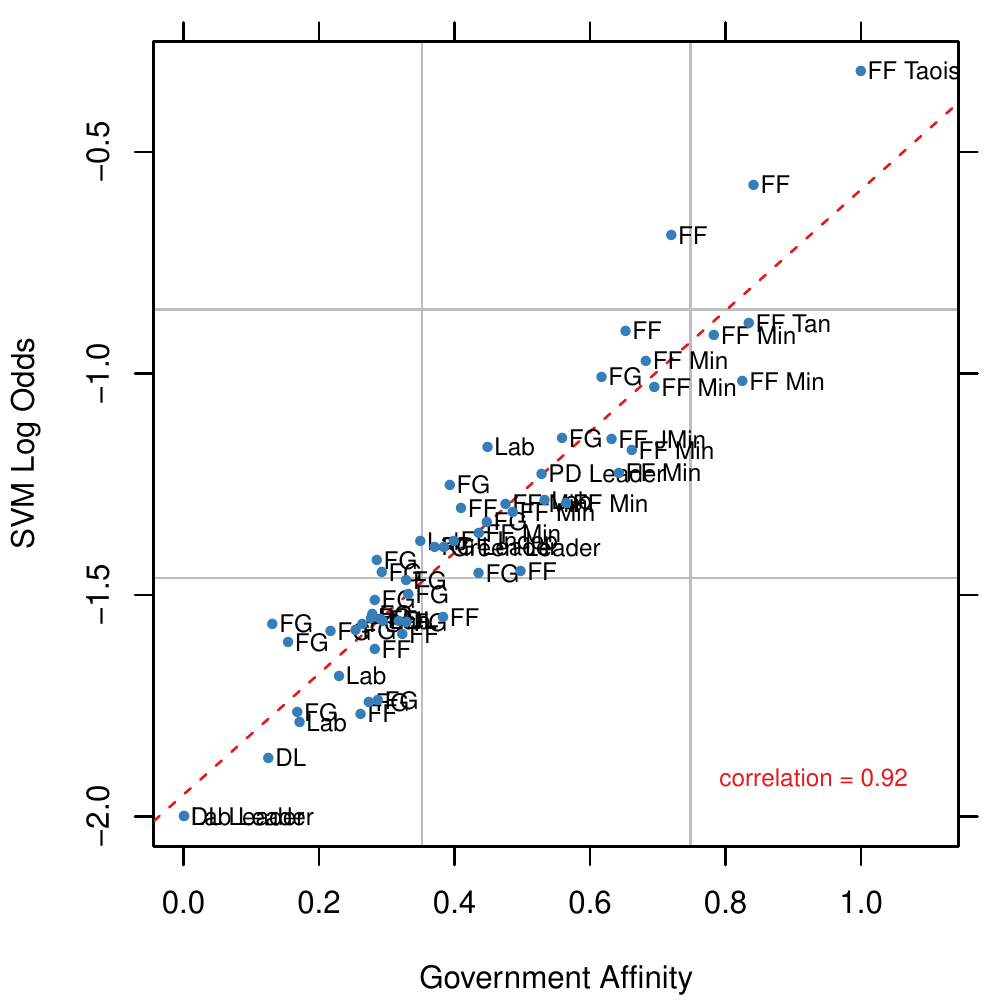}
  \caption{Support Vector Machine}
  \label{subfig:svm-corr}
 \end{subfigure}
 ~
 \begin{subfigure}[b]{0.425\textwidth}
  \includegraphics[width=\textwidth]{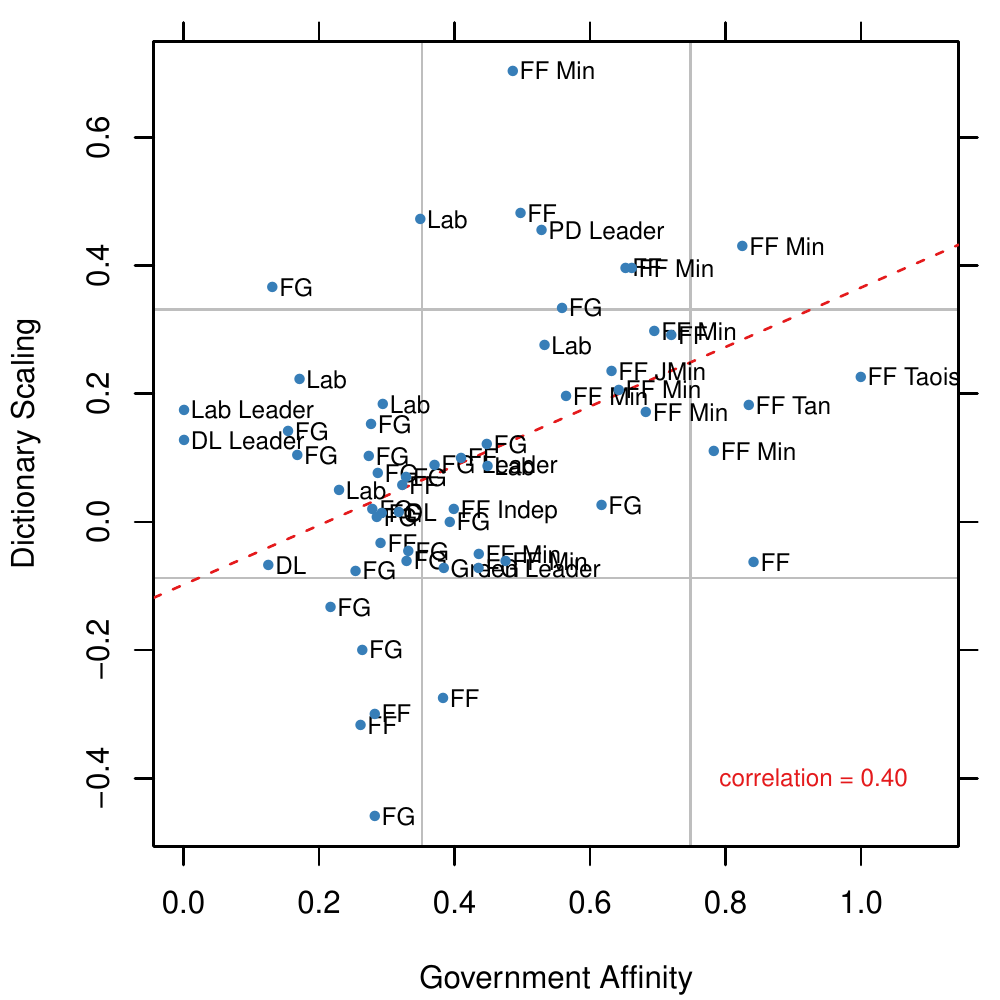}
  \caption{Dictionary}
  \label{subfig:dictionary-corr}
 \end{subfigure}
 ~\qquad
 \begin{subfigure}[b]{0.425\textwidth}
  \includegraphics[width=\textwidth]{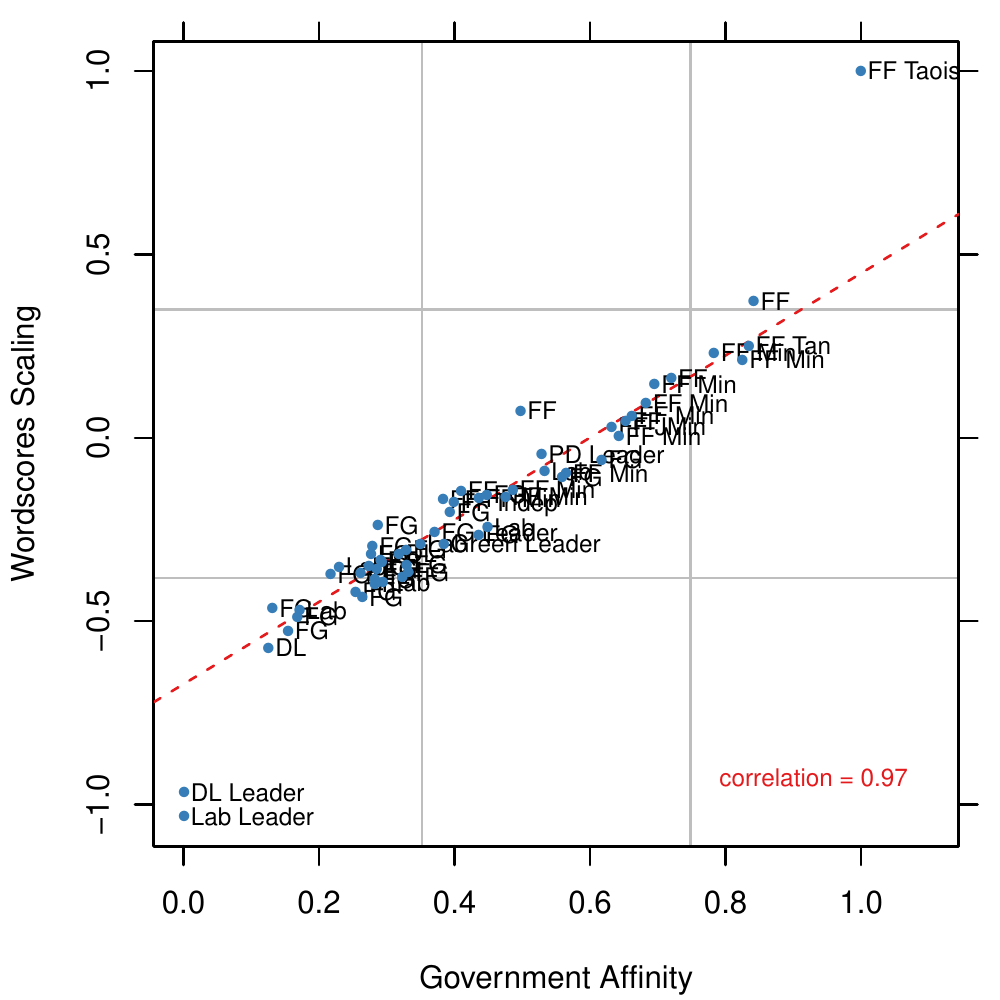}
  \caption{Wordscores}
  \label{subfig:wordscore-corr}
 \end{subfigure}
 ~
 \begin{subfigure}[b]{0.425\textwidth}
  \includegraphics[width=\textwidth]{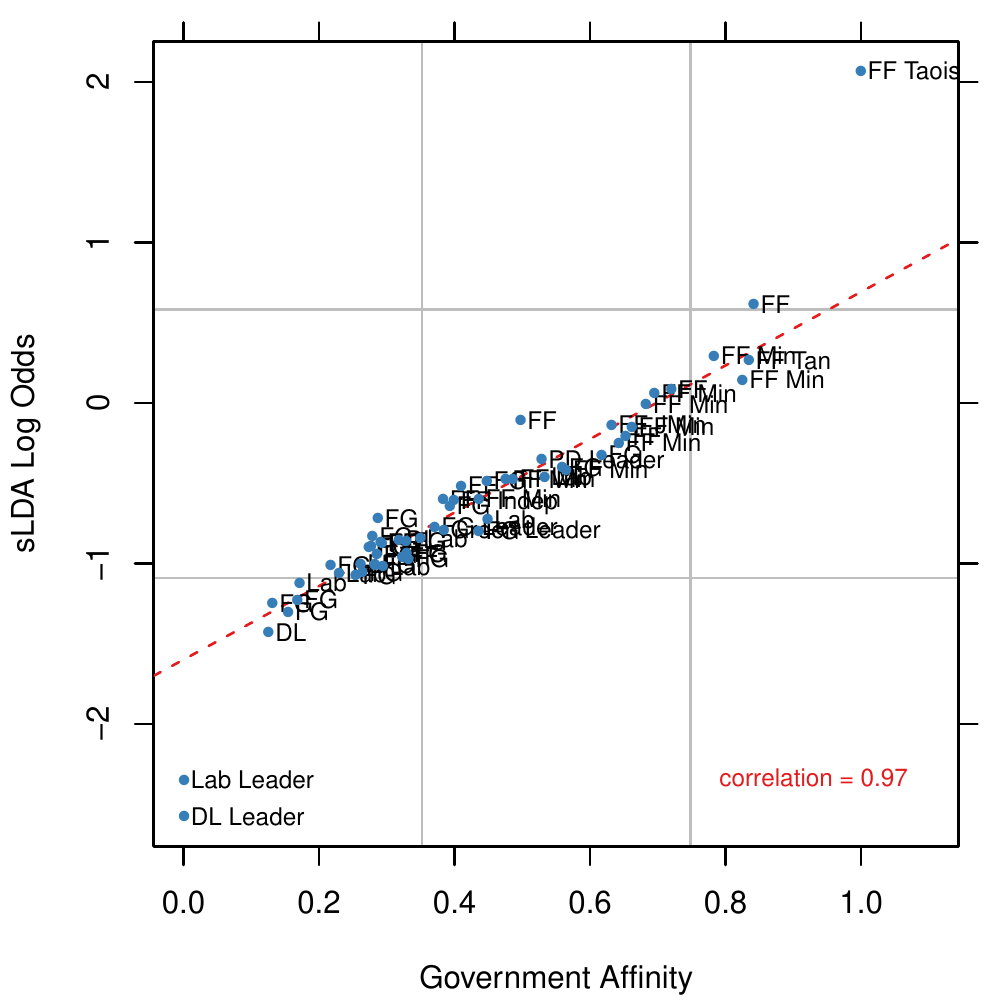}
  \caption{Supervised LDA}
  \label{subfig:slda-corr}
 \end{subfigure}
 ~\qquad
 \begin{subfigure}[b]{0.425\textwidth}
  \includegraphics[width=\textwidth]{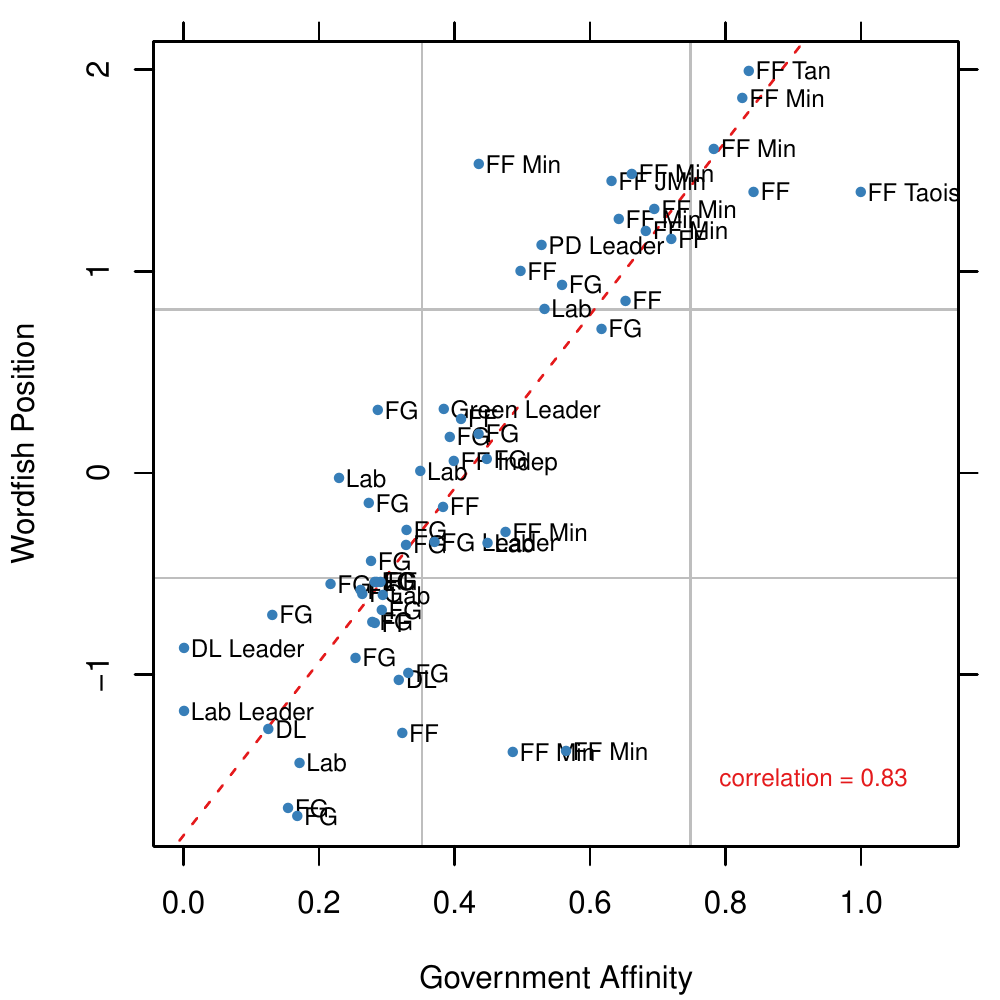}
  \caption{Wordfish}
  \label{subfig:wordfish-corr}
 \end{subfigure}
 \caption{Comparisons between scaling methods}\label{fig:scale-corr}
\end{figure}

\subsection{Dictionary methods}

In the special case that the reference distributions $p_1, p_2, \dotsc, p_K$
have disjoint supports---that is, when no two classes $k$ and
$l$ are such that both $p_k(v) > 0$ and $p_l(v) > 0$ for some word type
$v$---affinity scaling is exactly equivalent to dictionary scaling.

To make this equivalence clear, suppose that for each word type $v \in
\vocab$, at most one of the reference probabilities $p_{1v}, p_{2v}, \dotsc,
p_{Kv}$ is nonzero. When this is the case, we can partition the vocabulary as
a union of disjoint sets, $\vocab = \vocab_1 \cup \vocab_2 \cup \dotsb \cup
\vocab_K$, where
\[
 \vocab_k = \{ v \in \vocab : p_{kv} > 0 \}.
\]
Here, $\vocab_k$ is the set of word types associated with label $k$. The
disjoint support condition ensures that each word type $v$ is associated with
exactly one label.

Under the disjoint support condition, when we observe the $i$th token
$w_i$, we can
immediately infer the underlying orientation $u_i$ to be the only class with
this word in its support. The log-likelihood simplifies to
\begin{align*}
 l(\theta)
   & =
 \sum_{v \in \vocab}
 x_v \log \Big( \sum_{k=1}^{K} \theta_k p_{kv} \Big) \\
   & =
 \sum_{k = 1}^{K}
 \sum_{v \in \vocab_k}
 x_v \log (\theta_k p_{kv}) \\
   & =
 \sum_{k = 1}^{K}
 n_k \log \theta_k
 +
 \text{(constant)},
\end{align*}
where $n_k = \sum_{v \in \vocab_k} x_v$ and the constant does not depend on
$\theta$. In this case, the maximum likelihood estimate of the class affinity
vector is
\[
 \hat \theta =
 \left(
 \frac{n_1}{n},
 \frac{n_2}{n},
 \dotsc,
 \frac{n_K}{n}
 \right).
\]
That is, the estimated class affinities are the token occurrence rates in the
support sets $\vocab_1, \vocab_2, \dotsc, \vocab_K$.

\subsection{Wordscores}\label{sec:wordscores}

The ``Wordscores'' scaling method developed by~\citet{Laver+2003} turns out to
be closely related to class affinity scaling. That method, which is primarily
used to
scale documents between $K = 2$ reference classes works well in practice but
has been criticized for having \emph{ad hoc} theoretical
foundations \citep{Lowe2008}.  We can show, however, that Wordscores scaling is
closely related to affinity scaling, and gives highly correlated results for
texts that are not close to the extremes (represented by the reference text
positions).  We elaborate on this connection below.

In its simplest form, Wordscores takes as given
reference distributions for each class, denoted $p_1$ and $p_2$. The method
defines the wordscore of a word type $v \in \vocab$ as
\begin{equation}\label{eqn:wordscore}
 s_v =
 \frac{p_{2v} - p_{1v}}{p_{1v} + p_{2v}}.
\end{equation}
Word types that only appear in class~2 have scores of $+1$, while types that
only appear in class~1 have scores of $-1$. Other types have intermediate
values indicating the relative degrees of association with the two classes.
The unnormalized ``text score'' of a length-$n$ text with token count vector~$x$
is then the average wordscore of its tokens:
\begin{equation}\label{eqn:wordscores-scaling}
 t(x)
 =
 \frac{1}{n}
 \sum_{v \in \vocab} \frac{p_{2v} - p_{1v}}{p_{1v} + p_{2v}} x_v,
\end{equation}
Texts with positive $t(x)$ values tend to be more like class~2, while
texts with negative $t(x)$ values tend to be more like class~1.

The magnitude of the unnormalized score $t(x)$ is not directly
interpretable. To fix this,
\citet{MartinVanberg2007} advocate rescaling the score to ensure
that average reference texts from the two classes have scores of
$-1$ and $+1$. To realize the Martin--Vanberg scaling, for $k = 1,2$ define
\[
 t_k = \sum_{v \in \vocab} \frac{p_{2v} - p_{1v}}{p_{1v} + p_{2v}} p_{kv}.
\]
An average text of length $n$ from class~$k$ has token counts
satisfying $x_v/n = p_{kv}$, so that its score is $t(x) = t_k$.
Using the relation
\(
p_{1v}/(p_{1v} + p_{2v})
=
1 - p_{2v}/(p_{1v} + p_{2v})
\)
termwise in the sum, one can verify that $t_1 = -t_2$.
The Martin--Vanberg wordscore scaling is
\begin{align*}
 \tilde t(x)
   & =
 -
 \frac{t_2 + t_1}{t_2 - t_1}
 +
 t(x)
 \cdot
 \frac{2}{t_2 - t_1}
    = t(x) / t_2.
\end{align*}
An average text~$x$ from class~$1$ satisfies
$\tilde t(x) = -1$; an average text~$x'$ from class~$2$ satisfies
$\tilde t(x') = +1$.

The wordscore scaling $\tilde t(x)$ turns out to be deeply connected to
affinity scaling. To see this connection, note that using the parameterization
from Section~\ref{sec:estimating-affinities}, the score and observed information
functions for the affinity model evaluated at $\beta = 0$ are
\begin{align*}
 u(0)
   & =
 2
 \sum_{v \in \vocab}
 \frac{p_{2v} - p_{1v}}{p_{1v} + p_{2v}}
 x_v
 =
 2 n \, t(x),
 \\
 i(0)
   & =
 2 n
 \sum_{v \in \vocab}
 \frac{(p_{2v} - p_{1v})^2}{p_{1v} + p_{2v}}
 =
 2n \, (t_2 - t_1).
\end{align*}
There is a striking relationship between the scaled text score and
the derivatives of the mixture model log likelihood:
\[
 \tilde t(x) / 2 = \{i(0)\}^{-1} u(0).
\]
The right hand side of this expression is equal to the first Fisher scoring
iterate computed while maximizing $l(\beta)$ starting from the initial value
$\beta = 0$. When the maximizer $\hat \beta$ is close to $0$, it will
be approximately equal to this first iterate.  Thus, when a text is
roughly balanced between the two reference classes ($\hat \beta \approx 0$),
it will also be the case that
\[
 \tilde t(x)
 \approx
 2 \hat \beta
 =
 \hat \theta_2 - \hat \theta_1.
\]
For moderate documents, the wordscore scaling is a linear transformation of
the estimated class affinities.

We demonstrate the quality of this approximation in
Fig.~\ref{subfig:wordscore-corr}, where we plot the wordscore scaling versus
the estimated government affinity for the moderate debate speeches.
We can see that there is very good agreement between the two scalings, and
that $\tilde t(x) \approx 0$, the two scalings are almost identical.

\subsection{Support vector machines and logistic regression}

We have just shown analytically that affinity scaling gives similar results to
Wordscores. It turns out that, when the number of reference documents is
small, up to scaling, both methods are approximately equivalent to classifying
with a support vector machine or linear regression.

Suppose that we are in the two-class ($K = 2$) case, and that there is one
reference document for each class. Imagine fitting a linear classifier that
tries to predict class using a document's word frequencies as features. With a
vocabulary size $V$ greater than the number of training documents, the two
classes can be perfectly separated as long as the two reference distributions
$p_1$ and $p_2$ corresponding to the training documents are identical. In this
case, the support vector machine fit and the logistic regression fit are
identical, up to differences that arise from regularizing the coefficients.

Given a document with length $n$ and word count vector $x$, its feature vector
is its vector of word frequencies, $n^{-1} x$. The feature vectors for the two
training documents are $p_1$ and $p_2$. Up to a constant of proportionality,
the maximum margin predictor, expressed as a function of $x$ is
\begin{align}
 \eta(x) & = (p_2 - p_1)^\T \{ n^{-1} x - (1/2) (p_1 + p_2) \} \notag                \\
 \label{eqn:svm-scaling}
         & = \frac{1}{n} \sum_{v \in \vocab} (p_{2v} - p_{1v}) x_v + \text{(const.)}
\end{align}
Since the classes are perfectly separated, and multiple of this predictor
gives the same classification performance on the training set; the precise
scaling chosen by the fitting procedure will depend on the regularization
parameters.

Comparing the support vector machine scaling~\eqref{eqn:svm-scaling} with the
unnormalized wordscores scaling~\eqref{eqn:wordscores-scaling}, we can see
that the only substantive difference is the denominator $p_{1v} + p_{2v}$ in
the coefficient on $x_v$. Thus, up to a constant shift and scale, if
$p_{1v} + p_{2v}$ is roughly constant relative to $p_{2v} - p_{1v}$, then the
two methods will give similar results. In light of the connection between
Wordscores and affinity scaling developed in Sec.~\ref{sec:wordscores}, this
implies that in these situations, the support vector machine results will be
highly correlated with the affinity scaling results.

We verified the connection between the two methods empirically, using the
\texttt{SVMlight} software with the default tuning
parameters~\citep{Joachims1999}. Fig.~\ref{subfig:svm-corr} shows the support
vector machine estimated log odds plotted against the affinity scaling
results. Both scalings give similar results (correlation $0.92$). The main
distinction is that the numerical value of the support vector machine log odds
is determined completely by the regularization parameter and is thus
uninterpretable. The affinity scaling of a document, by contrast, can be
interpreted directly.

\subsection{Topic models}
\label{sec:topic-models}

Topic models share a similar perspective with the affinity model in that both
represent texts as mixtures of topics, with each topic having an associated
word distribution.  In our framework, the topics correspond to the reference
classes, and the text-specific topic weights correspond to class affinities.
We learn the class distributions from a set of labeled reference texts.  This
approach differs from that taken by unsupervised topic models
\citep{Blei+2003,Grimmer2010}, where estimated topics may or may
not correspond to scaling quantities of interest.

Supervised variants of topic models allow for associations between labels and
topics, but these models all assume that class membership is discrete, not a
continuous scale
\citep{McauliffeBlei2008,Ramage+2009,Roberts+2016}.  These supervised models
force clear associations between the topics and the scaling quantities of
interest, but they assume that the texts have discrete labels
indicating class membership. This fundamental assumption places these methods
in the same category as other classification methods like Naive Bayes,
estimating the probability of class membership, not class affinity.

Despite their philosophical differences, in practice supervised topic models
can give scalings that are highly correlated with the affinity model scaling.
The connection to supervised topic models is easiest to understand in the case
of \citepos{McauliffeBlei2008} Supervised Latent Dirichlet Allocation (sLDA),
which models a text-specific label as a random quantity linked to a linear
function of the text-specific topic weights.  Roughly speaking, the method
works in two stages. In the first stage, sLDA fits a topic model to the
reference texts. In the second stage, sLDA fits a logistic regression model
using the fitted topic weights as predictors and the class label as response.
In practice, sLDA fits the topics and the logistic regression simultaneously,
but when the number of topics is larger than the number of reference texts,
any differences between sequential and simultaneous fitting are determined by
the regularization parameters and the random initialization.

The connection between sLDA and affinity model scaling is closest with
two topics and two reference texts.  In this case, since the number of topics
equals the number of reference texts, sLDA can get a perfect fit by allocating
one topic to each reference text, and can separate the two classes perfectly
given the topic weights $(\hat \theta_1, \hat \theta_2)$ by using a
linear predictor for the odds of class membership of the form 
\(
 \eta = b (\hat \theta_2 - \hat \theta_1),
\)
where the coefficient $b$ gets determined by the regularization parameters.
When the sLDA fit gets used for prediction on
the unlabelled texts, the fitted topic weights $(\hat \theta_1, \hat \theta_2)$
will be the same as the values from a fitted affinity model (again, ignoring
the effects of regularization regularization and initialization).
The sLDA score will be highly correlated with the difference in estimated affinities.

In the case when there are more topics and more reference texts, the
relationship between affinity scaling and sLDA is not as simple, but the same
general intuition still holds and the two methods still give highly correlated
results. Fig.~\ref{subfig:slda-corr} illustrates this with a model using~10
topics, where the correlation between the non-reference text scalings from the
two methods is~$0.98$. Here, the sLDA method gives unreasonable results for
the extremes. Furthermore, the interpretation of the scaling value if different:
odds of class membership for sLDA, versus degree of membership for the
affinity model.

\subsection{Unsupervised methods}

Some approaches to scaling texts, including Latent Semantic Indexing
\citep{Deerwester+1990} and \cite{SlapinProksch2008}'s ``Wordfish'' Poisson
scaling method, estimate latent text-specific traits using unsupervised
methods.  Often, the estimated traits are correlated with recognizable
attributes, and so they can be used to scale ideology.
Letting $x_{iv}$ denote the count of word type~$v$ in text~$i$,
the \cite{SlapinProksch2008} Wordfish model specifies that
$x_{iv}$ is a Poisson random variable with mean $\lambda_{iv}$, where
\(
 \log \lambda_{iv}  = \alpha_i + \psi_v + \theta_i \, \beta_v
\)
for some unknown text-specific parameters ($\alpha_i$ and
$\theta_i$) and word-specific parameters ($\psi_v$ and $\beta_v$).
Estimates of $\theta_i$ have been shown to provide valid estimates of latent
positions expressed in speeches~\citep{LoweBenoit2013}.

The drawback to unsupervised scaling of this sort, however, is that they
provide  no guarantee that the estimated latent trait corresponds to the
quantity of interest. We demonstrate this behavior in
Fig.~\ref{subfig:wordfish-corr}, where we plot the Wordfish scaling estimates of
the debate speeches versus the affinity scaling estimates. The two methods give
similar results (correlation $0.82$), but there are also some notable
differences. The government and opposition leaders are not the most extreme
examples as determined by Wordfish, indicating  that even in this focused
context---a debate over a confidence motion---the  primary dimension
of difference is something other than the  government-opposition divide.


\section{Diagnostics}
\label{sec:diagnostics}

In the previous section, we used the simple analytic form of the affinity
scaling model to get an understanding of its
connections with other text scaling methods. Beyond this, we will now see
another advantage of the model's form: its simplicity facilitates
computationally efficient diagnostic checking for the model fit.

Ideally, our fit should exhibit two characteristics. First, it should not be
driven by a small number of word types, but instead it should be determined by
an accumulation of information from many different word types. Second, the
word types that show the most influence in determining the fit should be ones
that make sense from a subject matter perspective.  To check whether our
scaling results satisfy these properties, and to better understand them
generally, we will develop an influence measure to characterize the impact of
each word type in determining the overall fit.

Our strategy for assessing influence stems from \citet{Cook1977}, who,
in the context of linear regression, assesses the influence of each
observation by measuring the change that results from deleting the
observation. Proceeding analogously, we will measure the influence of a word
type $v \in \vocab$ by setting the corresponding token count $x_v$ to zero and
observing the change in the class affinity estimate $\hat \theta$.  Ideally,
we would do this by computing the maximizer $\hat \theta_{(v)}$ of the log
likelihood (or, when regularizing, the penalized log likelihood) gotten after
setting $x_v$ to zero, but the large number of word types makes this
impractical. We will settle for finding a computationally simple closed-form
approximation to $\hat \theta_{(v)}$.

Suppose that $x$ is a vector of token counts for the particular text of
interest, and that $\hat \theta = \theta_0 + C \hat \beta$ is the affinity
vector estimate gotten from $\hat \beta$, the maximizer of the
corresponding log likelihood $l(\beta)$ defined in~\eqref{eqn:affinity-loglik}.
Making the dependence on $x$ explicit, the score and observed information
functions are
\[
 u(\beta ; x)
 =
 C^\T Q x,
 \qquad
 I(\beta ; x)
 =
 C^\T Q X Q^\T C,
\]
where $X \in \reals^{\vocab \times \vocab}$ is a diagonal matrix with
$X_{vv} = x_v$ for $v \in \vocab$ and $Q = Q(\beta)$ is as defined in
Section~\ref{sec:estimating-affinities}.

For an arbitrary word type $v \in \vocab$, consider the effect of setting
$x_v = 0$. This defines a new vector of token counts $x_{(v)}$ defined by
$x_{(v)v} = 0$ and $x_{(v)w} = x_w$ for all $w \neq v$. Let $e_v$ denote
the $v$th standard basis vector in $\reals^{\vocab}$ and define
$h_v = C^\T \hat Q e_v$, where $\hat Q = Q(\hat \beta)$. Note that
$x = x_{(v)} + x_v e_v$, so that
  \[
 u(\hat \beta ; x)
   =
 u(\hat \beta ; x_{(v)}) + x_v \, h_v,
  \qquad
 I(\hat \beta; x)
   =
 I(\hat \beta; x_{(v)}) + x_v \, h_v h_v^\T.
\]
Since $u(\hat \beta ; x) = 0$, this implies that evaluating the score
function with the new data at the old estimate gives
\begin{equation}\label{eqn:u-delete-v}
 u(\hat \beta; x_{(v)})
 =
 -
 x_v \, h_v.
\end{equation}
The maximizer $\hat \beta_{(v)}$ of the new log likelihood is roughly equal to
the first Newton scoring step from $\hat \beta$. We can compute this
step explicitly by first computing the inverse of the observed information
matrix:
\begin{align}
 \{I(\hat \beta; x_{(v)})\}^{-1}
   & =
 \{I(\hat \beta; x) - x_v \, h_v h_v^\T \}^{-1}
 \notag
 \\
   & =
 \{I(\hat \beta; x)\}^{-1}
 +
 (x_v^{-1} - \tilde h_v^\T h_v)^{-1} \, \tilde h_v \tilde h_v^\T
 \label{eqn:iinv-delete-v}
\end{align}
where $\tilde h_v = \{I(\hat \beta; x)\}^{-1} h_v$.

Approximating the maximizer by the first Newton step from $\hat \beta$
gives
\begin{align*}
 \hat \beta_{(v)}
   & \approx
 \hat \beta
 +
 \{I(\hat \beta; x_{(v)})\}^{-1}
 \,
 u(\hat \beta; x_{(v)})
 \\
   & =
 \hat \beta
 -
 (x_v^{-1} - \tilde h_v^\T h_v)^{-1}
 \tilde h_v,
\end{align*}
where we have used~\eqref{eqn:u-delete-v} and~\eqref{eqn:iinv-delete-v} to
simplify the expression. Using this approximation for $\hat \beta_{(v)}$ gives
us an approximation for the change in the estimated affinities:
\begin{align*}
 \hat \theta
 -
 \hat \theta_{(v)}
   & =
 C \hat \beta
 -
 C \hat \beta_{(v)}
 \\
   & \approx
 (x_v^{-1} - \tilde h_v^\T h_v)^{-1}
 C \tilde h_v.
\end{align*}
Motivated by this approximation, we define our influence measure as
\begin{equation}\label{eqn:influence}
 d_v
 =
 (1/2)
 \| (x_v^{-1} - \tilde h_v^\T h_v)^{-1} C \tilde h_v \|_1,
\end{equation}
where $\|\cdot \|_1$ denoteds $1$-norm.
When we are regularizing the estimates, using a penalized log likelihood
$\tilde l(\beta; x)$ in place of $l(\beta; x)$, we define the influence
similarly, using the negative Hessian $- \nabla_\beta^2 \tilde l(\beta; x)$ in
place of $I(\beta; x)$.

Using a $1$-norm instead of a Euclidean norm in the definition of $d_v$
allows us to interpret $d_v$ as the total amount of positive change to the
components of $\hat \theta$. Given that
\(
1^\T ( \hat \theta - \hat \theta_{(v)} ) = 0,
\)
this is also equal to the total amount of negative change.


\section{Vocabulary selection}
\label{sec:vocabulary}

As previously mentioned, the results presented in Fig.~\ref{fig:scale-corr}
and elsewhere in the prequel use as vocabulary the set of word types appearing
in the leadership speeches, excluding words appearing only once and words
on the English Snowball ``stop'' word list. Why did we exclude these words?

Initially, we did not exclude any words from the vocabulary. We fit the
affinity model to the complete vocabulary and used it to scale the 55
non-leadership speeches. Then, to help understand our results, we computed the
influence measures as defined in~\eqref{eqn:influence} for each speech word
count vector~$x$ and word type~$v$. We also recorded the direction of the
influence (whether the appearance of the word pushes the fit towards
\Gov~or \Opp). This gave us a $55 \times 9731$
matrix of (speech, word) influence measures. Most of the entries of this
matrix are zero since most count vectors $x$ are sparse and words that do not
appear in a speech have no influence on its affinity estimate. For each word
type, we recorded the count of nonzero speech influence entries, along with
the median and maximum of the nonzero entries. We report these values in
Table~\ref{tab:influential-all}, grouped by the direction of influence.

\begin{table}
 \caption{Median and maximum influence ($\times 100$) exerted by
  the most influential words, grouped by direction of influence.
  Medians are computed over texts containing the word.
  }\label{tab:influential-all}
 \scriptsize

\begin{tabular}{lrrrclrrr}
\toprule
\multicolumn{4}{c}{Government} && \multicolumn{4}{c}{Opposition} \\
\cmidrule(r){1-4}
\cmidrule(l){6-9}
Word & Count & Median & Max && Word & Count & Median & Max \\
\midrule
and & 55 & 1.3 & 2.5 && the & 55 & 2.5 & 4.7 \\
our & 49 & 0.9 & 2.7 && that & 55 & 1.3 & 3.5 \\
graduate & 3 & 0.8 & 0.9 && to & 55 & 1.2 & 2.6 \\
deasy & 3 & 0.7 & 1.6 && they & 55 & 1.0 & 2.6 \\
attribute & 1 & 0.7 & 0.7 && a & 55 & 0.9 & 1.7 \\
social & 30 & 0.6 & 8.0 && is & 55 & 0.9 & 1.7 \\
per cent & 26 & 0.6 & 3.2 && not & 55 & 0.7 & 1.6 \\
corresponding & 1 & 0.6 & 0.6 && people & 54 & 0.7 & 3.0 \\
nation & 12 & 0.6 & 1.4 && it & 55 & 0.7 & 1.7 \\
proof & 2 & 0.6 & 1.0 && he & 42 & 0.6 & 2.0 \\
1987 & 20 & 0.5 & 2.7 && at & 54 & 0.5 & 1.3 \\
economic & 33 & 0.5 & 2.1 && his & 43 & 0.5 & 1.4 \\
will & 55 & 0.5 & 1.5 && taoiseach & 43 & 0.5 & 1.3 \\
international & 18 & 0.5 & 1.1 && by & 55 & 0.4 & 0.7 \\
union & 9 & 0.5 & 0.9 && as & 55 & 0.4 & 1.2 \\
\bottomrule
\end{tabular}

\end{table}

We can see, for example, that the word type \word{social} exhibited influence on
30 speeches. For one of these speeches, deleting the word \word{social} has the
affect of shifting the speech's affinity estimate away from
\Gov~by~$0.08$; the median shift for the 30~speeches is $0.006$. Deleting
\word{social} shifts the fit away from \Gov; equivalently, the
appearances of \word{social} push the fit towards~\Gov.

The influence of a word is determined by its usage rate and the degree to
which is usage is imbalanced across the reference classes.  The word types
that show up as influential in Table~\ref{tab:influential-all} are those that
appear frequently and exhibit a small imbalance between \Gov~and
\Opp, or else appear moderately and exhibit a large imbalance
between the two classes.  This holds generally: influential words tend to
either be highly imbalanced, or moderately imbalanced with high usage rates.

Many of the of the words in Table~\ref{tab:influential-all} make sense, for
example \word{social}, \word{nation}, and \word{economic} influence the
affinity fit
towards \Gov, and \word{people} and \word{taoiseach} influence the
affinity fit towards \Opp.  However, we can clearly see that
certain function words like \word{and} and \word{the} are exerting a big
influence
on the fit. These function words have slightly imbalanced usage rates in the
reference texts, which, compounded with a high usage rate, results in a large
net influence.  This sensitivity to stylistic differences is a manifestation
of a common critique of the related Wordscores scaling method
\citep{Beauchamp2012,GrimmerStewart2013}.  To reduce sensitivity to stylistic
differences, we eliminated function words (the Snowball English ``stop'' words)
from our analysis.

We can also see in Table~\ref{tab:influential-all} that there are words that a
few rare words like \word{attribute} and \word{proof} have large influence.
These
words are not meaningful discriminators on substantive grounds, but they show
up as influential because they only appear once in the reference speeches. The
estimated probabilities for these words are unreliable. Their influence is
determined purely by estimation variability. To get around this, in our final
analysis we choose to exclude these
words---the \emph{hapax legomena}---that only appear once in the reference
speeches.

\begin{table}
 \caption{Influential words after feature selection.
  Reporting is as described for Table~\ref{tab:influential}.
  }\label{tab:influential}
 \scriptsize

\begin{tabular}{lrrrclrrr}
\toprule
\multicolumn{4}{c}{Government} && \multicolumn{4}{c}{Opposition} \\
\cmidrule(r){1-4}
\cmidrule(l){6-9}
Word & Count & Median & Max && Word & Count & Median & Max \\
\midrule
deasy & 3 & 0.9 & 1.9 && people & 54 & 1.3 & 5.0 \\
per cent & 26 & 0.8 & 3.7 && taoiseach & 43 & 0.8 & 3.1 \\
nation & 12 & 0.8 & 1.8 && democrats & 23 & 0.7 & 1.9 \\
social & 30 & 0.8 & 10.7 && minister & 44 & 0.6 & 2.5 \\
corresponding & 1 & 0.7 & 0.7 && system & 37 & 0.6 & 2.7 \\
1990 & 17 & 0.7 & 2.0 && house & 54 & 0.5 & 1.9 \\
union & 9 & 0.7 & 1.0 && o'kennedy & 5 & 0.5 & 0.9 \\
belief & 3 & 0.7 & 1.0 && progressive & 24 & 0.5 & 1.4 \\
economic & 33 & 0.7 & 2.8 && say & 39 & 0.5 & 1.3 \\
reform & 19 & 0.7 & 2.4 && issue & 27 & 0.5 & 1.4 \\
1987 & 20 & 0.6 & 4.0 && million & 26 & 0.5 & 1.6 \\
policy & 27 & 0.6 & 2.0 && printed & 2 & 0.5 & 0.7 \\
roads & 6 & 0.6 & 2.6 && wealth & 6 & 0.5 & 1.4 \\
new & 38 & 0.6 & 1.6 && headings & 2 & 0.4 & 0.4 \\
international & 18 & 0.6 & 1.5 && said & 41 & 0.4 & 1.6 \\
\bottomrule
\end{tabular}

\end{table}

After excluding stop words and \emph{hapax legomena}, we were left with a
reduced vocabulary~$\vocab$ of 1321 word types.  We re-fit the model and
re-scaled the
speeches, computing the influences of the word types in the reduced-vocabulary
model. Table~\ref{tab:influential} shows the most influential
\Gov~and \Opp~words, computed as before.
It is possible that Snowball word list could have missed some influential
function words, but inspecting the words in Table~\ref{tab:influential} and
the other words further down in the order, we found that this was not the case
for our application. The only suspicious words are \word{say} and \word{said},
but in the context of the debate, it makes sense that these words are
pro-\Opp. When the word \word{said} gets used, it is typically used
to quote the government (``they said'' or ``they continue to say''), usually
by an opposition member criticizing the government. Likewise, at first glance
it may seem suspicious that \word{per cent} is at the top of the
\Gov~list, but in fact this often used to cite national
statistics about the economy and the GDP, using the state of the economy
explain the unrest.


\section{Uncertainty quantification}
\label{sec:uncertainty quantification}

In principle, it is possible to get standard errors for the affinity estimates
directly from  the expected or observed information
function~\eqref{eqn:affinity-oinfo}. However, these
likelihood-based standard errors are likely too narrow, because they ignore
uncertainty in the estimates of the reference distributions ($p_1, \dotsc,
p_K$), and they rely on the independence assumptions in the model. Ignoring
uncertainty in the reference distribution estimates is inappropriate when the
reference set is small, as it is here (three leadership speeches). Similarly,
the independence assumption---that word tokens in different positions of a
text are independent of each other---simplifies the analysis, but it is likely
violated in real-world data. To accurately assess the uncertainty in our
estimates, we need a method that accounts for the uncertainty in the
reference distribution estimates and the dependence between nearby words in
text.

To estimate the sampling distribution of the scaling estimates under dependence
between word tokens, we will use a block bootstrap that respects the natural
linguistic structure of the text, by following \cite{LoweBenoit2013}'s
recommendation to resample texts at the sentence level to  simulate sampling
variation but also to capture meaningful dependencies  among words within
natural syntactic units. To properly account for uncertainty in the reference
distribution estimates, we will also construct sentence-level bootstrapped
reference speeches. The full procedure is as follows:

\begin{enumerate}
 \item For bootstrap replicates $b = 1, \dotsc, B$:

       \begin{enumerate}
        \item For each reference text $y_1, \dotsc, y_R$ construct
              bootstrapped reference text $y_1^{\ast b}, \dotsc, y_{R}^{\ast b}$,
              where $y_i^{\ast b}$ has sentences drawn with replacement from
              $y_i$, with the same total number of sentences.

        \item Use the bootstrapped reference texts $y_1^{\ast b}, \dotsc,
              y_{R}^{\ast b}$ to estimate the reference distributions
              $\hat p_1^{\ast b}, \dotsc, \hat p_K^{\ast b}$ as
              described in Sec.~\ref{sec:estimating-ref-dists}.

        \item Construct a bootstrap version of the scaled text $x^{\ast b}$ by
              resampling sentences from $x$, with replacement.

        \item Treating the reference distribution estimates
              $\hat p_1^{\ast b}, \dotsc, \hat p_K^{\ast b}$ as fixed, construct
              an affinity-scaling estimate $\hat \theta^{\ast b}$ from $x^{\ast b}$.
       \end{enumerate}

 \item Use the sample standard deviation of $\hat \theta^{\ast 1}, \dotsc,
       \hat \theta^{\ast B}$ as the bootstrapped estimate of the standard
       error of the affinity scaling estimate $\hat \theta$ for $x$.

\end{enumerate}

We performed this procedure for all of 55 non-leadership speeches, getting a
separate bootstrap standard error for each. For comparison, we computed
likelihood-based (Wald) standard error for the estimates from the Fisher
information conditional on the reference estimates.
%
Unsurprisingly, the bootstrap standard errors are generally wider than the
likelihood-based estimates. The two uncertainty estimates are both on the same
order of magnitude, with the bootstrap standard error being less than 1.5
times as large as the likelihood-based standard error for most of the
speeches (87\%); the median ratio of the two standard errors is 1.3.
In the sequel, we use bootstrap standard errors to quantify the uncertainty
in the affinity estimates.

\begin{figure}
 \centering
 \includegraphics[width=5in]{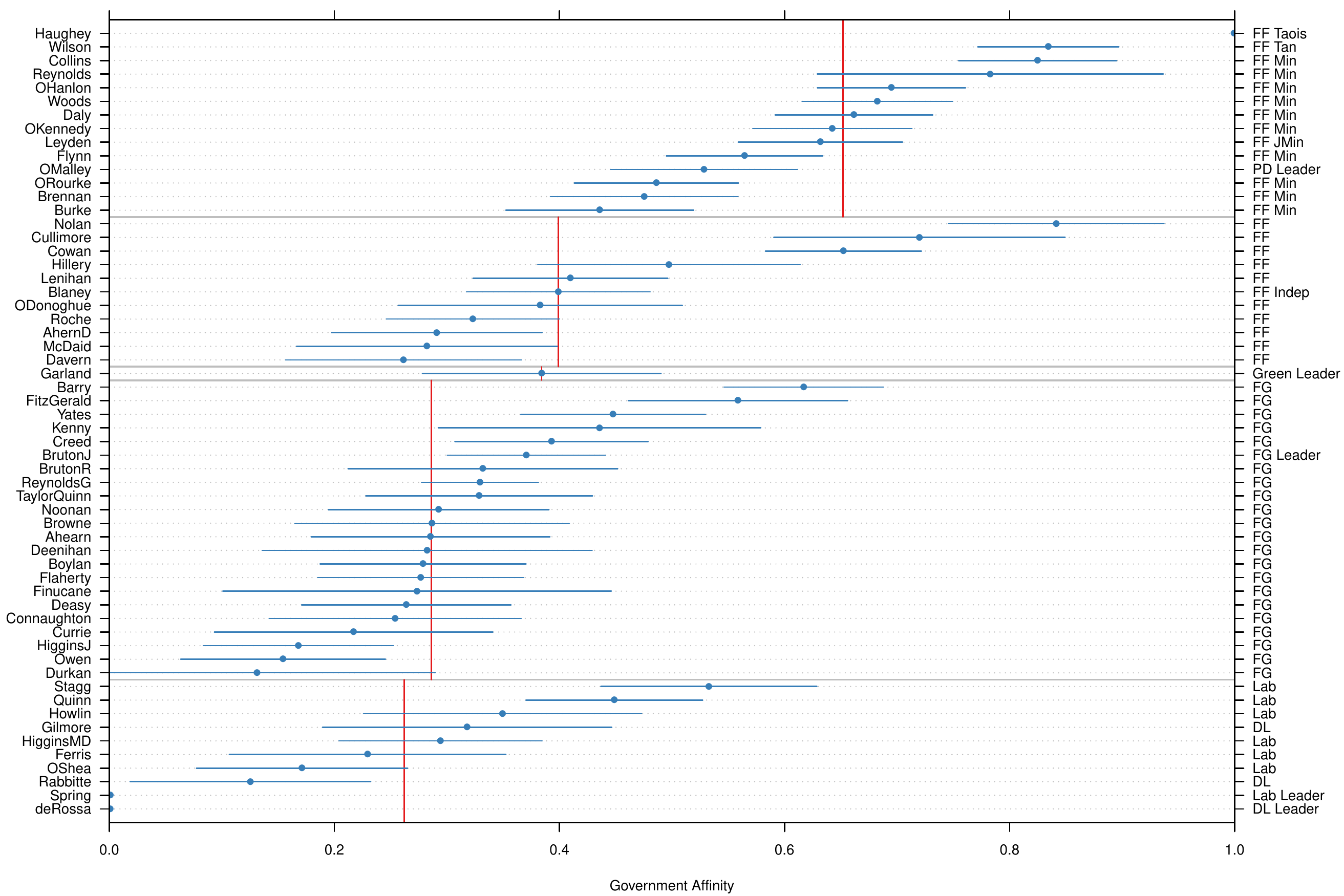}
 \caption{Affinity scaling estimates with bootstrap 95\% confidence intervals}
 \label{fig:ranking-boot}
\end{figure}

Fig.~\ref{fig:ranking-boot} displays the estimated government affinities for all
55 speeches after performing feature selection.  The figure includes 95\%
confidence intervals, computed using the sentence-level bootstrap. We discuss
these results in detail in the next section.

\section{Results}
\label{sec:results}


At both the level of the government versus opposition and inter-party levels,
the results are entirely in line with expectations: not only are the parties
arrayed in an order that would be consistent with expectations, with
opposition parties on the \Opp~side, and the governing parties on the
other, but also we see that speeches from the different parties align with the
extremity of their positions in regards to the establishment.   The speeches of
most centrist opposition party, \FineGael, express a more moderate
anti-\Gov~positions than either the left party Labour or the far-left
Democratic Left party.  This median difference emerges clearly even though we
considered the speeches of the Labour and Democratic Left leaders as
equivalent for the purposes of training the \Opp~class.

The more interesting distinctions emerge when we examine \emph{intra-}party
differences in expressed position. Among the government ministers, it is not
surprising to see that John Wilson, the FF Deputy Prime Minister
(\Tanaiste, or ``FF Tan'' in the plot), and
Gerard Collins, the Foreign Minister and a senior \FiannaFail\ minister
had extreme \Gov-oriented
estimated positions exceeded only by the \Taoiseach\ Charles
Haughey himself. What is more interesting is that the next minister in the
estimated ranking, Albert Reynolds, would later become the next
\Taoiseach. At the other extreme, among the most \Opp-oriented
government minister we see notable examples in Raphael (Ray) Burke,
who was removed from his ministerial position the following year,
and Mary O'Rourke, 
who months later would challenge Albert Reynolds for the party
leadership.


The ``back-bench'' FF members voted with the government but generally gave
speeches that were far more lukewarm than the FF ministers. Correspondingly,
we see that the estimated estimated \Gov~affinities for the
back-benchers are generally lower than those of the minsters. There were three
exceptions, members with extreme estimated \Gov-oriented affinities: Nolan,
Cullimore, and Cowan. One of these members, Brian Cowen, became Minister
for Labour the following year, and occupied senior positions include Prime
Minister for the next two decades.


On the opposition side, we see a similar set of heterogeneous estimated
affinities. Two salient examples of extreme estimated \Gov-oriented affinities
are \FineGael\ TD Garret FitzGerald, a former and future Prime Minister, and
TD Peter Barry, who had fought Fitzgerald in 1987 for party leadership. Both
emphasized fairly standard economic concerns, attacking the government's poor
economic performance rather than its corrupt behavior.
It is notable that the member with the highest estimated pro-opposition
affinity,
DL member Pat Rabbitte who would later become leader of the Labour Party;
in his speech, he
engaged in a personal set of attacks against the
\Taoiseach\ and specifically attacking his character and judgment.

The results of applying the class affinity scaling model to the confidence
debate speeches provides a results consistent with expectations and with
previous scholarly investigations of this episode \citep{LaverBenoit2002}.
Using only the texts of the speeches, we have succeeded at revealing
differences between the speakers that were not apparent from their party
affiliations.

\section{Discussion}

In our application and in others like it, the correct prediction of a class is
no longer a relevant benchmark because the process of producing political text
is expected to produce heterogeneous text within each class. For us, the
class---here, voting for or against the confidence motion, which was perfectly
correlated with government or opposition status---is observed and uninteresting,
while the heterogeneity is the primary interest. Despite what would seem obvious
from a measurement model or scaling perspective, however, a standard approach in
evaluating machine learning applications in political science has been
predictive accuracy benchmarked against known classes
\cite[e.g.][]{Evans+2007,Bei+2008}.
This focus on estimating correct classes not only wrongly shifts attention away
from the substantively interesting variation in latent traits, but also may
ultimately impair classification generality by encouraging over-fitting to
reduce predictive error.

Our proposed alternative, class affinity scaling, is based on a
probability model similar to those underlying class predictive methods, but
allows for mixed class membership.
We have shifted focus from
class prediction, something  typically uninteresting in the social sciences,
to a form of latent parameter estimation, while retaining
the advantages of supervised learning approaches where the analyst controls the
inputs that anchor the model.
While there is a strong tradition in some
disciplines, such as political science, of adapting machine learning to produce
continuous scales, practitioners are often unaware of
the differences in modeling assumptions between classification and scaling
methods \citep[e.g.][]{Laver+2003},
or they have not fully explored the implications of these assumptions
\citep[e.g.][]{Beauchamp2012}.  
We have highlighted the differences and similarities
in a form that encourages future development.

The relative simplicity of our method makes it amenable to direct mathematical
analysis. This simplicity allowed us to draw connections between Naive Bayes
classification, dictionary-based scaling, and a host of other methods.
We were further able to exploit the analytic simplicity of the
affinity scaling model to develop an influence measure assessing the
sensitivity of the fit, which we then used to
guide our vocabulary selection and to validate our fits to the \Dail~debate.


Using our method to explore the nuances of the speeches in the 1991
\Dail\ confidence motion, we produced estimates for each speaker that accord
with both
a qualitative reading of the speech transcripts and an expert understanding of
Irish politics.
Our application is a
hard domain problem, where no known lexicographical map exists to
differentiate government versus opposition speech and dictionary-based
scaling, even with a dictionary derived from political text, gives unsatisfactory results.
With limited training from the leadership speeches, class affinity scaling
is able to adapt to the context of the debate and give a meaningful scaling.
The method has
applications far beyond political text, however, and could be used to score more
standard sentiment problems on a continuous scale, or applied to any other
problem for which  contrasting reference texts can be identified.

\clearpage

\bibliographystyle{imsart-nameyear}
\bibliography{refs}


\end{document}